\documentclass{amsart}
\usepackage{graphicx}
\usepackage{amsmath}
\usepackage[a4paper, total={7in, 9in}]{geometry}
\usepackage{multirow} % Required for multirows
\usepackage{makecell}
\usepackage{xcolor}
\usepackage{enumerate}
\usepackage[hidelinks]{hyperref}

\date{December 2022}

\begin{document}
\title{Can Large Language Models design a Robot?}
\maketitle
Authors:

\author  *[1,2] {Francesco Stella,} \email{francesco.stella@epfl.ch,}

\author[2,3] {Cosimo Della Santina,} \email{c.dellasantina@tudelft.nl}

\author[1]  {Josie Hughes,} \email{josie.hughes@epfl.ch}\\

Affiliations: 

\address[1] {CREATE Lab, EPFL, Lausanne, Switzerland.} 

\address[2] {Department of Cognitive Robotics, Delft University of Technology, Delft, The Netherlands.} 

\address[3] {Institute of Robotics and Mechatronics, German Aerospace Center (DLR), Wessling, Germany.} 

*Corresponding author

\begin{abstract}\normalsize
 Large Language Models can lead researchers in the design of robots. 
\end{abstract}
Large Language Models (LLMs)  \cite{brants2007large}, are revolutionizing the field of robotics, providing robots with the ability to understand and process natural language at a level previously thought impossible. These powerful AI tools have the potential to improve a wide range of tasks in robotics, including natural language understanding, decision making, and human-robot interaction. One of the key advantages of large language models is their ability to process large amounts of text data, such as instructions, technical manuals, and maintenance logs, and internalize an implicit knowledge containing rich information about the world from which factual answers can be extracted. In fact, the text you have just read was generated by the LLM ChatGPT-3 \cite{floridi2020gpt} when prompted “Can you write an introduction in a newsy style to the potential of large language models in robotics?”. \\

Language models have long been used in robotics to translate natural language instructions into actions executable by robots \cite{ahn2022can}\cite{huang2022language}, synthesize code from text prompts \cite{budzianowski2019hello} and find relationships between different fields of knowledge. In light of these impressive capabilities, LLMs may now contribute to another bottleneck of robotics, design. Leveraging their emerging capabilities \cite{wei2022emergent}, LLMs can deliver a dialogue that enables, teaches, and guides humans in building a robot from scratch. These capabilities could fundamentally change the methodology by which we design robots, and could shift the role of humans from designer or engineer to technician. So, to what extent can ChatGPT-3 replace an engineer and design a robot?\\

\begin{figure}[tb]
\vspace{-0.25cm}
    \centering
    \includegraphics[width=1\linewidth]{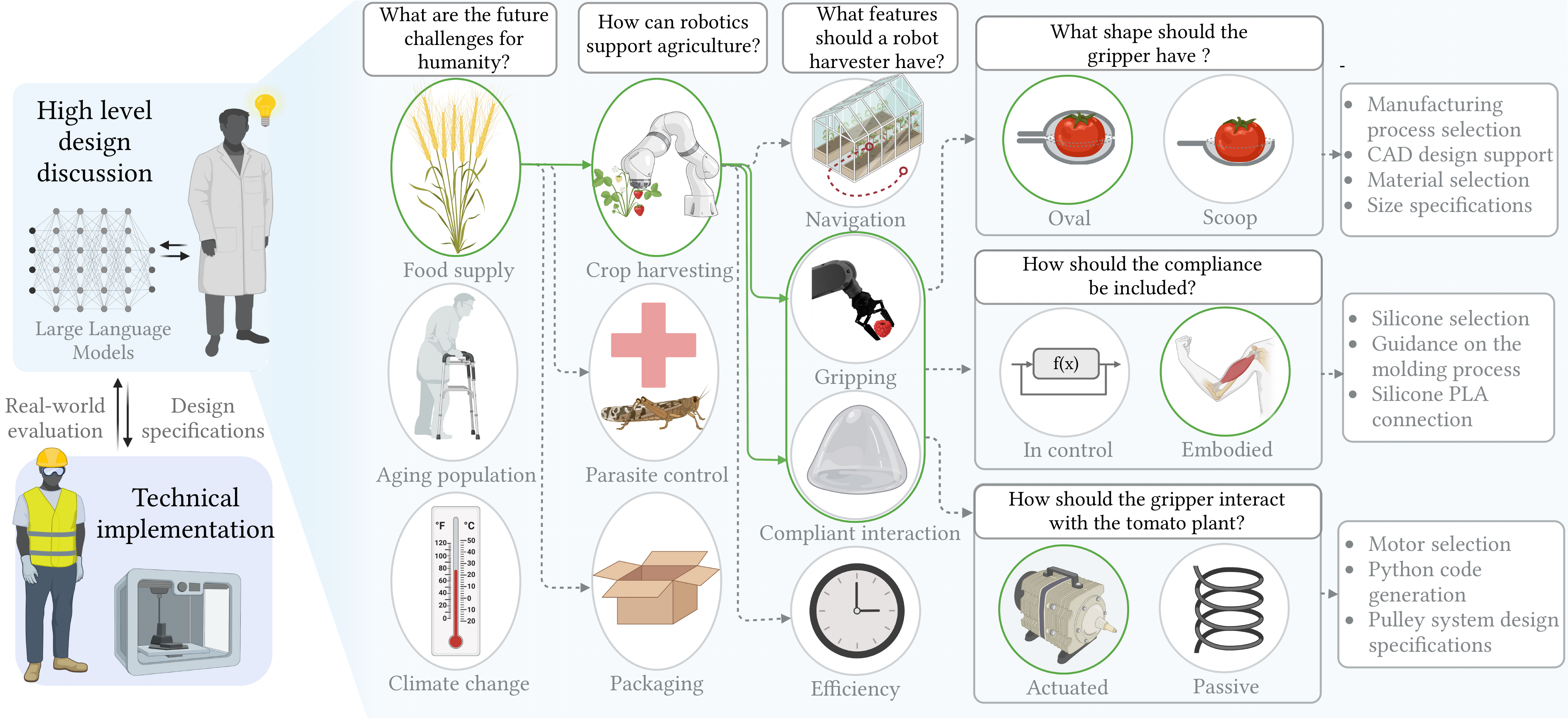}
%\vspace{-0.25cm}
    \caption{\small On the left, the two phases of the design process: first the human and LLM discuss the specifics application and of the design and later the human implements them. On the right, a pictorial overview of the discussion, with the questions prompted by the human on top, and the options provided by the LLM below. The green color highlights the decision tree of the human, which gradually focuses the problem to match his goal.}
    \label{fig:Concept}
   % \vspace{-1cm}
\end{figure}

To generate the first ChatGPT-3 designed robot we approach the task in a two step approach. In the first high-level phase, the computer and the human collaborate on a conceptual level, discussing ideas and outlining the specifications for the robot design while in the second phase the physical implementation of the design specifications takes place. 
As an example of this AI-driven design process, we consider the challenge of a human engineer driven by the desire to “help the world with robotics,” as shown in Figure \ref{fig:Concept}. The human operator starts by asking the LLM which are the future challenges for humanity and promptly gets an overview with a clear outline of the main hazards. The human can then select the option they are most interested in and narrow down the design space by asking for clarifications. This interaction can span multiple fields of knowledge and levels of abstraction, ranging from concepts to technical implementation. In this way, the human can spot new intersections between research fields, such as agriculture and robotics, and consider factors that are hardly part of the experience of an engineer by training, such as what is the crop that is economically most valuable to automate. By iterating this process, the LLM and the human converge to the technical design specifications of a robotic system.\\
  
\begin{figure}[ht]
\vspace{-0.25cm}
    \centering
    \includegraphics[width=\linewidth]{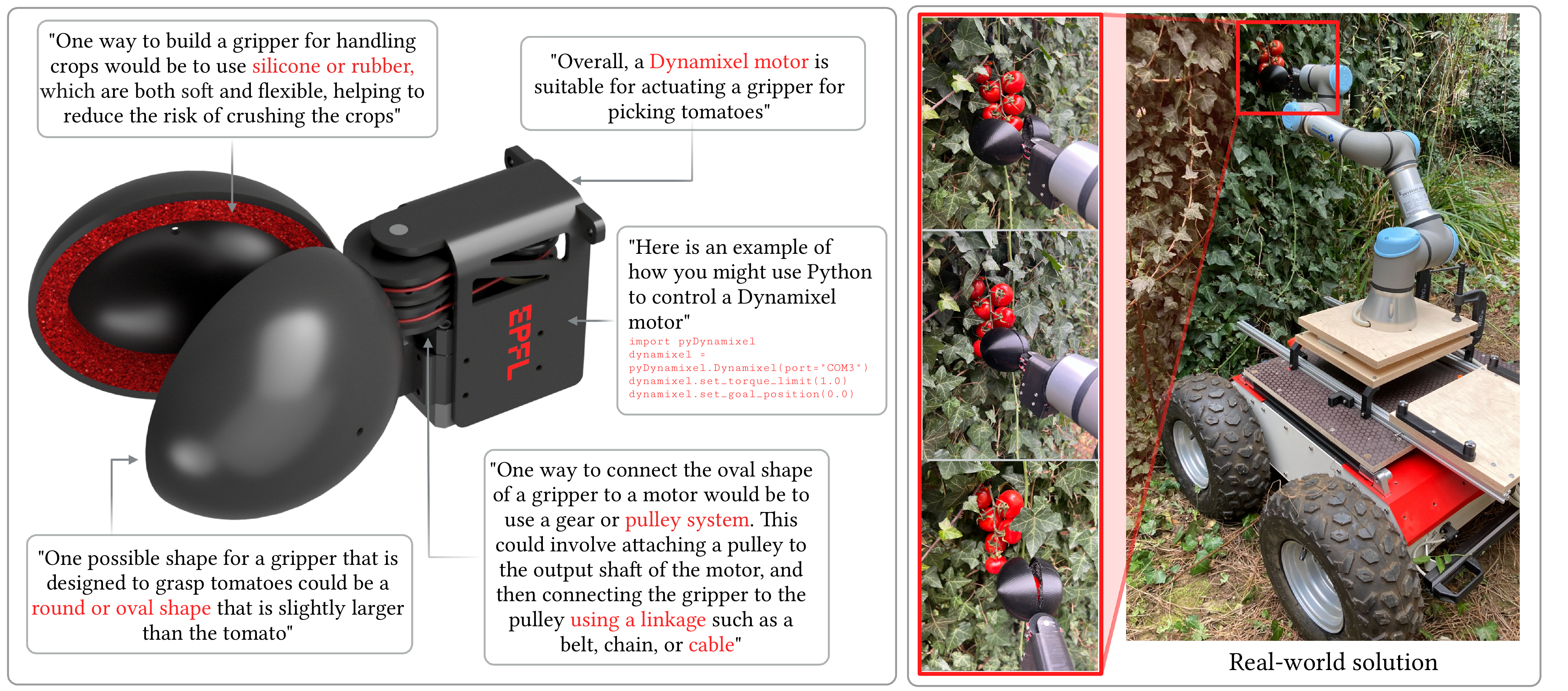}
%\vspace{-0.25cm}
    \caption{\small An AI designed this robotic gripper. }
    \label{fig:robot}
   % \vspace{-1cm}
\end{figure}

Typically, in a computational design framework, the computer solves technical problems specified by the human. In this case, conversely the LLM proposes conceptual options to the human, who then selects the most appealing choice. In this sense, the LLM acts as the researcher, leveraging knowledge and finding interdisciplinary connections, while the human acts as a manager, providing direction to the design. \
The application is selected as an output of this first part of the process, and a set of initial technical specifications are generated. This includes code, material, components, manufacturing method selection, and mechanism design. In the second, low-level phase of the design process, these directions need to be translated into a physical and functioning robot. Although LLMs can currently not generate entire CAD models, test code, or automatically fabricate the robot, recent advances have shown that AI algorithms can support the technical implementation of software \cite{chen2021evaluating}, mathematical reasoning \cite{Wolfram}, or even shape generation \cite{ramesh2022hierarchical}. Thus, we expect that, in the near future, AI-generated inputs will highly support a large set of technical tasks. However, in the foreseeable future, humans will remain mainly in charge of the technical implementation of the robotic solution. The human is therefore relegated to the technician role, polishing the code proposed by the LLM, finalizing the CAD, and fabricating the robot. This robot can then be tested in real-world scenarios, and a new conversation with the LLM can be used to iterate on the design process in light of experimental evidence. As an example of this second phase, Figure \ref{fig:robot} displays the main outputs generated by the LLM and the real-world deployment of the AI-designed robotic gripper for crop harvesting. \\

From this exploration we can foresee different modalities of human-AI interaction and collaboration. At one extreme, the LLMs could provide all the input required for robot design, which the human follows blindly. The AI is the inventor, addressing human questions and providing ‘creativity’, technical knowledge and expertise, whereas the human deals with the technical implementation. This could indirectly foster transfer and democratization of knowledge, by enabling non-specialists to realize robotic systems.  
A more moderate, yet powerful approach is collaborative exploration between the LLM and the human, leveraging the ability of the LLM to provide interdisciplinary and wide ranging knowledge to augment the human's expertise. Finally, we can consider a third approach in which the LLM acts as a funnel, helping to refine the design process and providing technical input whilst the human remains the inventor or scientist involved in the process. This collaboration between AI and humans presents clear benefits and opportunities. By augmenting human knowledge with LLMs, this methodology removes the limits imposed by the learning process and supports the human in finding relevant connections between fields, making interdisciplinary research and reasoning more accessible. It can spur the curiosity of researchers, interactively teach new robotics engineers, and accelerate the design process. As seen in our demonstration, the relationship between human and AI may vary for different parts of the design process depending on the skill and expertise of the individual and the goal of the robotic design process.\
At the same time, the introduction of LLMs into the design of robots brings questions regarding its potentially negative effects on scientific disciplines and engineering - a creative, interdisciplinary, and IP-creating process that currently relies on highly-skilled professionals. 
In this regard, it is pivotal to point out that LLMs should be regarded as an evolution of search engines, generating the "most probable" answer to a given prompt \cite{shanahan2022talking}. As such, it is debatable if they can develop creative solutions that substantially advance the robotics discipline beyond what is already known by the scientific community. But unlike search engines, LLMs can propose ways to integrate 'knowledge' and apply it to unseen problems, thus providing a potentially false impression that new knowledge is being generated.
Moreover, we see another potential issue in the widespread use of LLMs in our field. As the same trained model is accessible to everybody, it could create a bias in researchers' focus toward solutions that the model statistically prefers. This way, it may hinder the exploration of new technological solutions.
Finally, this ability of the LLM to apply and adapt prior experience to new problems could prevent humans from taking responsibility for the solutions developed \cite{stokelchatgpt}, which could lead to dangerous outcomes and a lack of human creativity in the design process. This could prohibit and stagnate the advancement of new robotic technologies and designs. There are also significant societal and ethical implications resulting from human-AI interactions for robot design. LLMs could automate high-level cognitive design tasks, and have humans focusing on more technical jobs. This could redefine the set of skills that are required by an engineer, and change the education that engineers should receive. 
 Finally, there are key issues regarding plagiarism, traceability and IP \cite{USPATENT}.  Can a design created via LLM be considered to be novel as it builds only on prior knowledge \cite{george2022can}, and also how can this previous knowledge be referenced? Similarly, if human-AI collaboration leads to the creation of novel IP, is this not a function of the training data of the LLM?  As this technology matures there are also longer term considerations including data-privacy, the frequency of retraining and how new knowledge should be integrated to maintain the usability and relevancy of this tool.\\
 
To conclude, the robotics community must identify how to leverage these powerful tools to accelerate advances and capabilities, yet doing so in an ethical, sustainable and socially empowering way. We must develop means of acknowledging the use of LLMs \cite{lee2023can}, and also being able to trace the lineage of the generation of designs from LLM.  Looking forward, we strongly believe that LLMs will open many exciting possibilities and that they will be a force for good if opportunely managed. The design process could be fully automated by combining collaborating LLMs to ask and answer questions, with one helping to refine the other. This could also be augmented with automated fabrication to allow for a fully autonomous pipeline for the creation of bespoke and optimized robotic systems. Ultimately, it is an open question for the future of this field if these tools can be used to assist robot developers and leverage inter-disciplinary knowledge leading to new robotic capabilities, or does this lead to a long-term stagnation of the field, with lazy, unskilled engineers, relying on external computation to generate new knowledge?

\bibliographystyle{naturemag}

\bibliography{References}
\iffalse
\clearpage
\fi

\end{document}